%% file: main.tex
\documentclass[journal, twoside]{IEEEtran}

\usepackage{acro}
\input{acronym}

\usepackage{siunitx}

\usepackage{datetime2}

\usepackage{comment}
\usepackage{xspace}
\usepackage{booktabs}
\usepackage{gensymb}
\usepackage{algorithmic}
\usepackage{orcidlink}

\usepackage{tikz}
\usetikzlibrary{arrows.meta, shapes.geometric, positioning, fit, backgrounds}
\definecolor{primary}{HTML}{D81B60} 
\definecolor{secondary}{HTML}{FFC107}
\definecolor{tertiary}{HTML}{1E88E5}

\usepackage{amsmath}

\newcommand{\OS}{\overline{OS}}

\ifCLASSOPTIONcompsoc
  \usepackage[caption=false,font=normalsize,labelfont=sf,textfont=sf]{subfig}
\else
  \usepackage[caption=false,font=footnotesize]{subfig}
\fi

\begin{document}

\title{Navigation by Skylight Polarization via Rayleigh Model Optimization on a Road Vehicle}

\author{Ben Potter\orcidlink{0009-0005-4972-7897}, Eslam Mounier\orcidlink{0000-0002-7020-2842}, Aboelmagd Noureldin\orcidlink{0000-0001-6614-7783}, Yahia Antar, and Muhammad Alam\thanks{Corresponding author: Muhammad Alam}\thanks{Ben Potter is with the Department of Electrical and Computer Engineering, Queen’s University, Kingston, ON K7L 3N6, Canada (e-mail: ben.potter@queensu.ca).}\thanks{Eslam Mounier is with the Department of Electrical and Computer Engineering, Queen’s University, Kingston, ON K7L 3N6, Canada, and also with the Department of Computer and Systems Engineering, Ain Shams University, Cairo 11535, Egypt (e-mail: eslam.abdelmoneem@queensu.ca).}\thanks{Aboelmagd Noureldin is with the Department of Electrical and Computer Engineering, Royal Military College of Canada, Kingston, ON K7K 7B4, Canada, and also with the School of Computing, Queen’s University, Kingston, ON K7L 2N8, Canada (e-mail: aboelmagd.noureldin@rmc.ca).}\thanks{Y. M. M. Antar is with the Electrical Engineering Department, Royal Military College of Canada, Kingston, ON K7K 7B4 Canada (e-mail: antar-y@rmc.ca).}\thanks{Muhammad Alam is with the Department of Electrical and Computer Engineering, Queen’s University, Kingston, ON K7L 3N6, Canada (e-mail: m.alam@queensu.ca).}}

\IEEEtitleabstractindextext{%
\begin{abstract}
Reliable absolute heading estimation is critical for road-vehicle navigation, particularly when magnetic, inertial, or GNSS-based references are degraded. 
This paper presents a skylight-polarization navigation method that estimates vehicle heading by globally optimizing a Rayleigh sky model against measurements from a division-of-focal-plane polarimetric camera. 
The proposed framework combines a pinhole camera model, a single-scattering Rayleigh polarization model, and a solar position algorithm to synthesize the expected full-image angle-of-polarization distribution for candidate vehicle states. 
A degree-of-polarization-weighted root-mean-square error objective function compares measured and simulated polarization patterns, allowing the estimator to use all available sky information while attenuating low-confidence pixels caused by clouds, buildings, foliage, and other depolarizing occlusions. 
Experimental evaluation is performed using three urban road-vehicle trajectories collected with a roof-mounted polarimetric camera and a high-precision dual-antenna GNSS/INS reference system. 
Results show sub-degree RMS yaw error across all trajectories, ranging from \qty{0.718}{\degree} to \qty{0.889}{\degree}. 
Compared with representative feature-based and model-based polarization navigation methods, the proposed approach reduces RMS heading error by up to \qty{87}{\percent} and demonstrates improved robustness to partial sky occlusion.
\end{abstract}

}

\maketitle

\IEEEdisplaynontitleabstractindextext
\IEEEpeerreviewmaketitle

\section{Introduction} \label{sec:introduction}
\input{sections/introduction.tex}

\section{Related Work} \label{sec:related_work}
\input{sections/related_work.tex}

\section{Theory of Skylight Polarization for Navigation} \label{sec:theory}
\input{sections/theory.tex}

\section{Model Optimization} \label{sec:model_optimization}

\input{sections/model_optimization.tex}

\section{Instrumentation and Dataset} \label{sec:dataset_instrumentation}
\input{sections/dataset_instrumentation.tex}

\section{Results} \label{sec:results}
\input{sections/results.tex}

\section{Conclusion} \label{sec:conclusion}
\input{sections/conclusion.tex}

\IEEEtriggeratref{36}
\bibliographystyle{IEEEtran}
\bibliography{IEEEabrv,references}

\end{document}

%% file: acronym.tex
\DeclareAcroArticle{definite}{the}
\DeclareAcroEnding{possessive}{'s}{'s}

\NewAcroCommand\dac{m}{\acrodefinite\UseAcroTemplate{first}{#1}}
\NewAcroCommand\Dac{m}{\acroupper\acrodefinite\UseAcroTemplate{first}{#1}}

\NewAcroCommand\Dacg{m}{%
  \acroupper
  \acrodefinite
  \acropossessive
  \UseAcroTemplate{first}{#1}%
}

\DeclareAcronym{pcm}{
  short = PCM,
  long  = pinhole camera model,
}

\DeclareAcronym{pso}{
  short = PSO,
  long  = particle swarm optimization,
}

\DeclareAcronym{svm}{
  short = SVM,
  long  = support vector machine,
}

\DeclareAcronym{gnss}{
  short = GNSS,
  long  = global navigation satellite system,
}

\DeclareAcronym{wrmse}{
  short = WRMSE,
  long  = weighted root mean squared error,
}

\DeclareAcronym{rmse}{
  short = RMSE,
  long  = root mean squared error,
}

\DeclareAcronym{aop}{
  short = AoP,
  long  = angle of polarization,
}

\DeclareAcronym{dop}{
  short = DoP,
  long  = degree of polarization,
}

\DeclareAcronym{doa}{
  short = DoA,
  long  = division-of-aperature,
}

\DeclareAcronym{dofp}{
  short = DoFP,
  long  = division-of-focal-plane,
}

\DeclareAcronym{ncp}{
  short = NCP,
  long  = north celestial pole,
}

\DeclareAcronym{enu}{
  short = ENU,
  long  = East-North-Up,
}

\DeclareAcronym{imu}{
  short = IMU,
  long  = inertial measurement unit,
}

\DeclareAcronym{ppp}{
  short = PPP,
  long  = precise point positioning,
}

\DeclareAcronym{rmc}{
  short           = RMC,
  long            = Royal Military College of Canada,
  short-definite  = \nospace ,
  long-possessive = '
}

\DeclareAcronym{psa}{
  short = PSA,
  long  = Plataforma Solar de Almeria
}

\DeclareAcronym{ht}{
  short = HT,
  long  = Hough transform
}

%% file: sections/introduction.tex
\IEEEPARstart{M}{odern} mobile platforms spanning driver assistance systems, manned tactical vehicles, autonomous road transportation, and industrial robotics depend on accurate, continuous, and highly resilient navigation systems for safe and reliable operation. 
Among the core parameters governing vehicle state estimation and trajectory control, absolute heading determination plays a pivotal role. 
In dynamic, real-world operational environments, maintaining an uncompromised orientation reference is essential to prevent cumulative dead-reckoning drift, ensuring operational continuity and human safety, particularly when external positioning infrastructure is degraded or unavailable.

Despite significant advances in navigation instrumentation, conventional sensors utilized for absolute heading determination suffer from severe physical and environmental limitations. 
Gyrocompasses and \acp{imu} provide self-contained dead-reckoning capabilities, but their orientation estimates suffer from unbounded error integration over time~\cite{rogne2014ImuDrift, jimenez2010ImuDrift, georgy2010ImuDrift}, and a loss of power permanently destroys the reference orientation. 
Magnetic compasses provide absolute heading but are highly susceptible to local electromagnetic interference~\cite{fu2021MagneticInterference, jianchengFang2011MagneticCalibration, yan2016MagneticCalibration}, structural ferromagnetic distortion from the vehicle chassis~\cite{vasconcelos2011MagneticCalibration, chung2011LocalizationMagnetic}, and ambient anomalies in the Earth's magnetic field~\cite{kao2006MagneticCalibration, li2013MagneticReview}. 
Dual-antenna \ac{gnss} configurations can deliver high-precision heading solutions under open skies; however, their performance degrades in complex operational environments. 
Natural and artificial obstructions such as tall buildings and dense foliage cause signal multipath and line-of-sight shadowing~\cite{ollander2018GnssMultipath, he2011GnssMultipath, georgiadou1988GnssMultipath}.
Furthermore, \ac{gnss} signals are increasingly vulnerable to deliberate radio frequency jamming and spoofing~\cite{mehr2025GpsJamming, grant2009GpsJamming}, which poses failure risks in high-reliability civilian and defense scenarios.
Although multi-sensor fusion strategies aim to combine complementary sensor modalities to mitigate individual weaknesses, effective fusion relies on the availability of a passive, drift-free, and infrastructure-independent absolute orientation reference.

Skylight polarization offers a compelling physical foundation for self-contained, passive, and drift-free navigation. 
As unpolarized sunlight enters the Earth's atmosphere, atmospheric scattering interactions impart a predictable, structured linear polarization pattern across the sky. 
The spatial distribution of this polarization pattern depends solely on the solar position relative to an observer on the Earth's surface.
Biological organisms, including desert ants~\cite{wehner2006AntCompass, duelli1973PolarizedOrientation}, honeybees~\cite{kraft2011HoneybeeNavigation, vonFrisch1965BeeOrientation}, desert locusts~\cite{zittrell2020LocustCompass, pegel2017LocustCompass}, and monarch butterflies~\cite{reppert2004PolarizedMonarch}, exploit this skylight polarization pattern via specialized optical structures to perform long-distance navigation with remarkable precision~\cite{horvath2014PolarizationVision}.
Inspired by these biological visual systems, polarization-based navigation systems measure skylight polarization parameters across observed sky points to infer the absolute orientation of the observer.

Recent developments in polarimetric imaging, particularly \ac{dofp} polarimeters, have enabled real-time, high-resolution capture of spatial polarization patterns.
Traditional polarization navigation algorithms predominantly rely on extracting specific geometric landmarks, such as the solar meridian line or neutral points, using line-fitting or transformation techniques such as the \ac{ht}~\cite{lu2015HoughTransform}.
However, these local feature-based methods suffer from three primary drawbacks when applied to ground vehicle navigation.
In urban canyons, buildings, bridges, and tree canopies obscure significant portions of the sky. When local geometric features (e.g., the solar meridian) are occluded, feature-based algorithms experience performance degradation or failure.
Existing studies often assume a strictly horizontal polarimeter orientation. In mobile vehicle applications, dynamic pitch and roll motions introduce non-horizontal measurement tilts. Transforming \ac{aop} measurements from the sensor frame into ray coordinate frames under non-horizontal conditions incurs systematic errors that are rarely modeled or compensated in literature.

To overcome these challenges, this paper presents a global model optimization framework for absolute heading determination on road vehicles using a wide-field \ac{dofp} polarimetric camera. 
Rather than relying on local feature extraction, our method formulates heading estimation as an optimization problem evaluated over the full polarimetric image. 
By integrating a \ac{pcm} with single-scattering Rayleigh sky theory and an accurate solar position algorithm (\ac{psa})~\cite{blancoMuriel2001SolarVector, blanco2020SolarVector}, we simulate superpixel-level expected \ac{aop} distributions for candidate vehicle poses. 
We construct an objective function that compares simulated and measured \ac{aop} distributions, weighted by the \ac{dop}, across all visible sky pixels. 
The \ac{dop} weighting naturally attenuates low-polarization noise from depolarizing cloud cover and solid occlusions.

The main contributions of this work are summarized as follows:
\begin{enumerate}
  \item 
    We develop a full-image polarimetric camera simulation combining a \ac{pcm} with the Rayleigh sky model to synthesize expected superpixel-level \ac{aop} distributions for \ac{dofp} sensor hardware under arbitrary camera orientations and solar positions.
    Additionally, we formulate our simulation based exclusively on the state (i.e., position, time, orientation) of a virtual polarization camera in the horizontal coordinate system, which enables its application to conventional trajectory data where the position of the sun is not immediately recorded.
  \item 
    We formulate a coordinate transformation between the \ac{dofp} measurement coordinate system and individual ray coordinate systems used by the Rayleigh sky model.
    We explicitly account for non-horizontal camera tilt and document the errors incurred from such tilted measurements in vehicle-mounted schemes.
  \item 
    We propose a \ac{wrmse} objective function for heading extraction that evaluates global skylight polarization patterns and incorporates \ac{dop} weighting to suppress measurement noise and depolarizing occlusions such as cloud cover, buildings, and foliage.
  \item We perform experimental evaluations on an urban vehicle dataset collected under dynamic road conditions using a high-precision dual-antenna \ac{gnss}/\ac{imu} reference system. Experimental results demonstrate that our method achieves up to an \qty{87}{\percent} decrease in the \ac{rmse} of the heading estimate compared with traditional \ac{ht}-based methods and up to a \qty{73}{\percent} decrease compared with other model optimization methods.
\end{enumerate}

The remainder of this paper is structured as follows. 
Section~\ref{sec:related_work} provides a review of relevant literature on bio-inspired skylight polarization navigation. 
Section~\ref{sec:theory} presents the physical theory of skylight polarization, representations of linearly polarized light, and the Rayleigh sky model.
Section~\ref{sec:model_optimization} details the proposed model optimization framework, objective function formulation, camera projection model, and polarimetric measurement scheme. 
Section~\ref{sec:dataset_instrumentation} describes the experimental vehicle instrumentation and the urban trajectory dataset.
Section~\ref{sec:results} presents the experimental results, performance evaluations, and comparative analyses against existing methods.
Finally, Section~\ref{sec:conclusion} concludes the paper and highlights directions for future research.

%% file: sections/related_work.tex
Methods for heading estimation from polarized skylight measurements fall into two major categories.
The first category relies on the extraction of heading from geometric landmarks, such as the solar merdian and neutral points, present in images of the skylight polarization.
The \ac{ht} was introduced as a method of line-fitting the solar meridian by Lu and collaborators~\cite{lu2015HoughTransform}, and has been implemented on diverse hardware~\cite{li2025PolHeadingHetro}.
Other works have used the \ac{ht} method as a benchmark for their datasets.
Unsupervised learning methods, such as the \ac{svm}, have been suggested for solar merdian detection, too~\cite{liang2020SVMHeading}.
The neutral points have been used for full attitude determination by Jue et al.~\cite{jue2026SimulatedAnnealing}.
Such geometric landmark identifiers suffer from performance degradation when regions of the polarization measurement are occluded.
While these methods show promise on static datasets collected with rotational stages, they begin to suffer under the dynamic conditions of realistic road trajectories with occlusions from buildings, foliage, and adverse weather.

The second category uses an analytic model of the skylight polarization pattern to resolve the heading estimate.
Wang et al. established the application of the orthogonality of the e-vector and solar vector predicted by the Rayleigh model using the Lagrange multiplier method to determine the solar vector given a set of measured e-vectors~\cite{wang2017BionicOrientation}.
Kronland-Martinet et al. use the distribution of the \ac{dop} to determine heading~\cite{km2025NorthPole}.
Zhou et al. use the additivity of the e-vectors in circles about the zenith~\cite{zhou2025AdditivityPolarizationPattern}.
He et al. apply the Berry model to optimize the solar vector using the \ac{pso} algorithm~\cite{he2025BerryModelOptimization}.
While such methods have explored sky model optimization as a technique for heading determination, additional research is required for improved performance in realistic scenarios. 

Model-based methods for preprocessing measured polarization data have also received significant research attention.
Optical path modelling, via frameworks like the \ac{pcm}, has been used for calibration~\cite{qin2025PolarizedCameraModel, zhang2023ModelOrientationDetermination} and noise suppression~\cite{li2025WaveInterferenceSuppression}.
Other methods use a sky model to resolve the \ac{aop} in occluded regions of the polarization measurement~\cite{li2026PatchRestoration, jue2026SimulatedAnnealing, fan2025OrientationRayleigh}.
Typically, such restoration of the \ac{aop} data is followed by the application of an existing heading extraction method: \cite{fan2025OrientationRayleigh} uses the Lagrange method from~\cite{wang2017BionicOrientation} and \cite{li2026PatchRestoration} uses the \ac{ht} method from~\cite{lu2015HoughTransform}.
Our proposed method combines the strength of the sky model under occluded settings with the heading estimation algorithm directly, which reduces the importance of such preprocessing schemes.

%% file: sections/theory.tex
\subsection{Coordinate System Definitions}

\subsubsection{Horizontal Coordinate System}

The horizontal coordinate system, denoted the $n$-frame, has axes $E$, $N$, and $U$, with an origin located at the observer, $O$.
Following the convention for horizontal coordinate systems, its axes are aligned with the cardinal east and north directions with the $U$ axis pointing upward.
All other coordinate systems are placed relative to the $n$-frame.

\subsubsection{Camera Body Coordinate System}

The camera body coordinate system, denoted the $b$-frame, has axes $X^b$, $Y^b$, and $Z^b$, with an origin located at the observer, $O$.
Since the origins of the $n$-frame and $b$-frame are colocated at the observer, they only differ in rotation, not translation.
The rotation between these two frames is the orientation of the camera.
Our method aims to constrain the component of this rotation which corresponds to an angle about $U$.
This angle is the yaw, also referred to as the heading of the camera.

\subsubsection{Ray Coordinate System}

The ray coordinate system, denoted the $r$-frame, has axes $X^r$, $Y^r$, and $Z^r$, where its origin and orientation in the $n$-frame depend on the direction of propagation of the ray.
Indeed, rays propagating towards the observer from different directions have different transformations between their respective $r$-frame and the $n$-frame.
Given a propagation direction defined by the position of the sky point, $P$, which is the origin of the ray, and with respect to the observer $O$, which is the destination of the ray, the $Z^{r}$ axis points along the direction of propagation from $P$ to $O$, the $Y^{r}$ axis is orthogonal to $Z^r$ in the $EN$ plane, and the $X^{r}$ axis completes the right-handed coordinate system.

\subsubsection{Superpixel Coordinate System}

The superpixel coordinate system, denoted the $s$-frame, has axes $X^s$, $Y^s$, and $Z^s$ and its origin depends on the position of the superpixel on the image sensor.
The $s$-frame is defined relative to the $b$-frame, since each superpixel is physically embedded in the camera hardware.
Its axes point in the same direction as the $b$-frame, but they differ in their origin.
We define the coordinate system of the $i^{th}$ superpixel as $X^s_i$, $Y^s_i$, and $Z^s_i$.

\subsection{Representations of Linearly Polarized Light}

For linearly polarized light, we represent the orientation of the e-vector as an angle defined in the plane which is transverse to the direction of propagation.
This angle is commonly referred to as the \ac{aop}.
However, light is rarely perfectly coherent (i.e., organized at a single \ac{aop}) and is usually a mixture of polarized and unpolarized components.
We represent the ratio between the polarized and unpolarized components by a fraction of total intensity referred to as the \ac{dop}.
The \ac{aop} and \ac{dop} represent the relevant state of the linearly polarized ray of light.
We only consider linearly polarized light since other works have established that elliptical polarization states due to atmospheric scattering are negligible~\cite{cronin2011PolarizedLight}.

We derive the \ac{aop} in the $r$-frame, denoted $a^r$, given the e-vector, denoted $\overrightarrow{E^r}$, in (\ref{e:aop_from_e}).
We take the \ac{aop} from the $X^r$ axis in the right-handed rotation direction.
We note that the \ac{aop} for $\overrightarrow{E}$ is identical to the \ac{aop} of $-\overrightarrow{E}$ due to (\ref{e:aop_from_e}).
Indeed, negation of the e-vector does not change the state of polarization since its physical significance is to define the plane of oscillation which is symmetric under negation.
Therefore, we typically express the \ac{aop} on the interval $(-\frac{\pi}{2}, \frac{\pi}{2}]$.
Fig.~\ref{f:scatter_event} shows the construction of $a^r$, given $\overrightarrow{E^r}$ in the $r$-frame.

\begin{equation} \label{e:aop_from_e}
  \tan a^r = \frac{ E^r_y }{ E^r_x }
\end{equation}

\subsection{Rayleigh Model}

The single-scattering Rayleigh model predicts the \ac{aop} and \ac{dop} of the light that gets scattered towards the observer from an observed sky point based on its geometry in relation to the sun.
Such a scattering event occurs when light from the sun interacts with a particle in the atmosphere causing it to become polarized.
Before the scattering event, the ray is unpolarized (i.e., $\acs{dop} = 0$), as it has propagated directly from the sun.
After the scattering event, the ray becomes polarized and the state of polarization is estimated by the Rayleigh model.

In such single-scattering models, we assume the ray only undergoes one of such interactions before reaching the observer.
Certain features of the skylight polarization pattern, such as the neutral points, are only present when multiple scattering effects are modelled.
As such, other sky models have been proposed, which vary in fidelity and fall into two major categories: solving the vector radiative transfer equation and analytic descriptions.
Solving the vector radiative transfer equation provides high accuracy, but its computational cost is prohibitive in navigation applications.
Notable analytic descriptions of the skylight polarization pattern include the Rayleigh model and the Berry model~\cite{berry2004BerryModel}.
The main benefit of the Berry model over the Rayleigh model is its ability to model the neutral points.
However, we implement our method with the Rayleigh sky model as it does not depend on accurate modelling of the neutral points, it is the most common sky model used in the existing literature, and prior work has shown that it matches measurements remarkably well~\cite{suhai2004RayleighPerformance}.

\begin{figure}[!t]
    \centering
    \subfloat[]{%
      \includegraphics[width=1.74in]{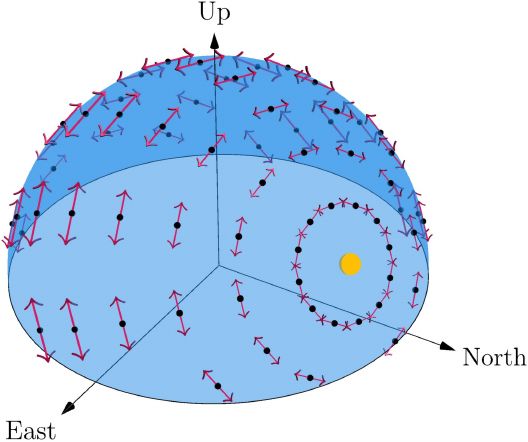}%
      \label{f:evectors}%
    }
    \subfloat[]{%
      \includegraphics[width=1.74in]{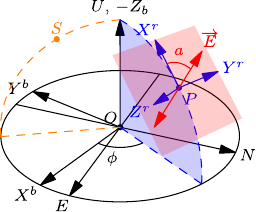}%
      \label{f:scatter_event}%
    }
    \caption{Rayleigh-model geometry for skylight polarization. (a) Predicted e-vector orientations over the sky form a structured pattern around the solar point. (b) A single-scattering event defines the scattered ray observed at $P$ and the corresponding e-vector orientation orthogonal to the scattering plane.}
    \label{f:rayleigh}
\end{figure}

In Fig.~\ref{f:evectors}, we show the predicted distribution of the e-vector for many sky points which are uniformly distributed about the solar sky point.
We see that the Rayleigh model predicts a ring-like distribution of e-vector orientations that occur orthogonally from the direction towards the sun.
This characteristic distribution of e-vectors, thus states of polarization, is the basis of the skylight polarization pattern.

We consider a single-scattering event depicted in Fig.~\ref{f:scatter_event} that occurs at the point $P$ by a ray traveling from the sun at point $S$.
The Rayleigh model predicts that the e-vector of the ray after the scattering event is orthogonal to the plane defined by $O$, $P$, and $S$.
We use the cross product in (\ref{e:e_n}) to define the e-vector in the $n$-frame using the vectors from $O$ to $S$ and $O$ to $P$.
The e-vector is then transformed from the $n$-frame to the $r$-frame by (\ref{e:e_r}) using the matrix $C^r_n$ where we can determine the \ac{aop} using (\ref{e:aop_from_e}).

\begin{align} 
  \overrightarrow{E^n} &= \overrightarrow{OS} \times \overrightarrow{OP} \label{e:e_n}\\
  \overrightarrow{E^r} &= C^r_n \overrightarrow{E^n} \label{e:e_r}
\end{align}

The \ac{dop} of the ray is defined in terms of the scattering angle, $\gamma$, which is the angle at $O$ between $S$ and $P$.
We define the scattering angle in (\ref{e:gamma}).
Then, we use the scattering angle to determine the \ac{dop}, denoted $d$, in (\ref{e:dop_from_gamma}).
The \ac{dop} is unitless and does not require a reference direction like the \ac{aop}.

\begin{align}
  \cos \gamma &= \overrightarrow{OS} \cdot \overrightarrow{OP} \label{e:gamma} \\
  d      &= \frac{ \sin^2 \gamma }{ 1 + \cos^2 \gamma } \label{e:dop_from_gamma}
\end{align}

We augment the Rayleigh sky model with a solar position algorithm to enable the computation of the skylight polarization pattern directly from the position and time of the observer.
We determine the position of the sun, $\OS$, using the \ac{psa} algorithm~\cite{blancoMuriel2001SolarVector}.
The \ac{psa} algorithm computes the direction of the solar vector based on the location and time at the observer.
We use the location and time provided by the reference system as inputs to the algorithm.
The algorithm is popular for open-loop computation of the solar vector due to its balance between accuracy and computational complexity.
When it was originally published, the algorithm was accurate to \qty{30}{\arcsecond} (\qty{0.00833}{\degree}) for the years 1999 through 2015.
More recently, the original authors have updated the algorithm for the years 2020 through 2050~\cite{blanco2020SolarVector}.

\begin{figure}[!t]
    \centering
    \subfloat[Measured AoP in Superpixel Frame]{%
      \includegraphics[width=1.74in]{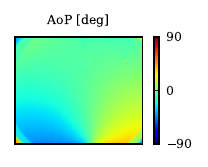}%
      \label{fig:imgs_aop_sensor}%
    }\hfil
    \subfloat[Measured AoP in Ray Frame]{%
      \includegraphics[width=1.74in]{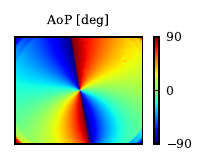}%
      \label{fig:imgs_aop_global}%
    }\\
    \subfloat[Measured DoP]{%
      \includegraphics[width=1.74in]{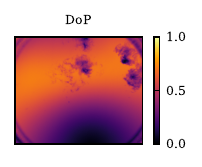}%
      \label{fig:imgs_dop}%
    }\hfil
    \subfloat[Simulated AoP in Ray Frame]{%
      \includegraphics[width=1.74in]{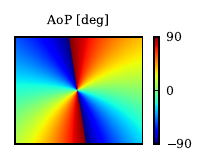}%
      \label{fig:imgs_sim_aop}%
    }
    \caption{Example measured and simulated polarization images used by the proposed model-optimization framework. The measured \ac{aop} is shown in the superpixel frame and after transformation to the ray frame, alongside the measured \ac{dop} and the Rayleigh-model \ac{aop} prediction generated from the reference state.}
    \label{fig:imgs}
\end{figure}

We show false-color polarization visualizations in Fig.~\ref{fig:imgs}.
Fig.~\ref{fig:imgs_aop_sensor},~\ref{fig:imgs_aop_global},~and~\ref{fig:imgs_dop} show measured polarization data captured during our data collection campaign.
Fig.~\ref{fig:imgs_aop_sensor} shows the \ac{aop} in the $s$-frame (i.e., as it is recorded by the \ac{dofp} polarimeter).
Fig.~\ref{fig:imgs_aop_global} shows the \ac{aop} in the $r$-frame, after the superpixel to ray frame transformation is applied.
Fig.~\ref{fig:imgs_dop} shows the \ac{dop} of the sky.
Sparse cloud cover can be seen in the \ac{dop} distribution, since water clouds are known to have a depolarizing effect.
Finally, Fig.~\ref{fig:imgs_sim_aop} shows the model predicted \ac{aop} distribution in the ray coordinate system which is simulated using the state provided by the reference system.
We note that the \ac{aop} distribution in Fig.~\ref{fig:imgs_aop_global} and Fig.~\ref{fig:imgs_sim_aop} match closely.
The similarity between these two distributions is the key constraint on the model optimization algorithm.

%% file: sections/model_optimization.tex
We propose a method of absolute heading estimation from skylight polarization based on optimization of the Rayleigh sky model.
Our method combines a \ac{pcm} with the Rayleigh model to simulate images of the skylight polarization pattern given the state (i.e., position, time, and orientation) of a virtual camera.
We define an objective function that encodes the proximity of an estimated state of a virtual camera to the true state of a given measured image using this simulation.

\begin{figure}[t]
    \centering
    
    \tikzset{
        block/.style = {rectangle, rounded corners, minimum width=2.0cm, minimum height=1cm, text centered, align=center, font=\small, draw=black, fill=white},
        groupbox/.style = {draw=gray, dashed, thick, fill=gray!10, rounded corners, inner sep=10pt},
        arrow/.style = {thick, ->, >=stealth}
    }

    \begin{tikzpicture}[node distance=0.5cm and 0.5cm]

    \node (cameramodel) [block] {Camera\\Model};
    \node (skymodel) [block, right=of cameramodel] {Rayleigh\\Sky Model};
    \node (objective) [block, right=1cm of skymodel] {Objective\\Function};

    \node (orientation) [block, above=1.5cm of cameramodel] {Orientation};
    \node (postime) [block, above=1.5cm of skymodel] {Position\\and Time};

    \node (measurement) [block, above=of objective] {Measurement};

    \draw [arrow] (orientation) -- (cameramodel);
    \draw [arrow] (postime) -- (skymodel);
    \draw [arrow] (cameramodel) -- (skymodel);
    \draw [arrow] (skymodel) -- (objective);
    \draw [arrow] (measurement) -- (objective);

    \begin{scope}[on background layer]
        \node [groupbox, 
               fit=(cameramodel) (skymodel), 
               label={[font=\bfseries]above:Simulation}] (simulationbox) {};
    \end{scope}

    \begin{scope}[on background layer]
        \node [groupbox, 
               fit=(orientation) (postime), 
               label={[font=\bfseries]above:State Estimate}] (statebox) {};
    \end{scope}

    \end{tikzpicture}
    
    \caption{High-level flow for evaluating the model-optimization objective function. Candidate orientation, position, and time are used to simulate the expected skylight polarization image, which is then compared with the measured polarimetric image.}
    \label{f:flow}
\end{figure}
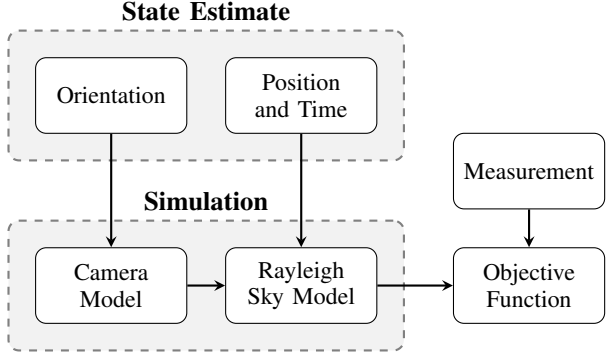

Fig.~\ref{f:flow} provides a high-level overview of the evaluation of the objective function for a given state estimate and measurement.
The camera model uses the orientation of the state estimate to map each pixel onto a corresponding sky point.
The sky model uses the position and time of the state estimate to determine the position of the sun.
Then, it computes the predicted state of polarization for each mapped sky point from the camera model.
Finally, the objective function evaluates the difference between the predicted \ac{aop} of each pixel against real measurements.

\subsection{Objective Function}

The objective function is defined in~(\ref{equ:j}).
It is derived as a \ac{wrmse} between the measured \ac{aop}, $a^r$, and the expected \ac{aop}, $\hat{a}^r(\mathbf{s})$, given the state $\mathbf{s}$.
The state is composed of a position, orientation, and time.
We weight the error in the expectation by the measured \ac{dop} to increase the invariance of the objective function to noisy \ac{aop} measurements.
The objective function combines the weighted expectation error for each pixel in the measured image.
We assign each pixel an index, $i$, and denote the measured values for the $i$th pixel in the subscript.

\begin{equation} \label{equ:j}
  J(\mathbf{s}) = \sqrt{ \frac{ \sum d_i [a^r_i - \hat{ a }^r_i (\mathbf{s})]^2 }{ N \sum d_i } }
\end{equation}

The expected \ac{aop} is computed using the Rayleigh model and a camera model.
We consider each pixel in a polarization image to be an independent measurement of the polarization state of a ray originating at a unique point in the sky.
This point is determined by the pixel's location on the image sensor and the optical properties of the camera.
By building a model of the camera, we can map pixel locations onto unique sky points.
Then, we apply this geometry to find the expected \ac{aop} using the Rayleigh model.
In priciple, generating the expected \ac{aop} values is the same as simulating a virtual polarization camera.

\subsection{Pinhole Camera Model}

We implement a geometrical model for tracing the path of a light ray from a sensor pixel onto a sky point.
The $Z^{b}$ axis points away from the viewing direction and is commonly referred to as the optical axis, principal axis, or principal ray.
The $X^{b}$ and $Y^{b}$ axes form a plane called the principal plane.
At the origin of the camera coordinate system, denoted $O$, is the \emph{aperture}.
The \emph{image plane} is defined by the image coordinate system which is 2-dimensional and centered at $R$ with basis vectors $U$ and $V$.
All light from the scene, destined for the image plane, must pass through the aperture.
We constrain the position of $R$ by the focal length, denoted $f$, such that $R = fZ^b$, where $f \geq 0$.
By changing $f$, we control the field of view through which the \ac{pcm} views the scene.
We found that a focal length of \qty{8}{\milli\metre} matched our lens hardware well, but this can be tuned if desired.

\begin{figure}[t]
  \centering
  \includegraphics[]{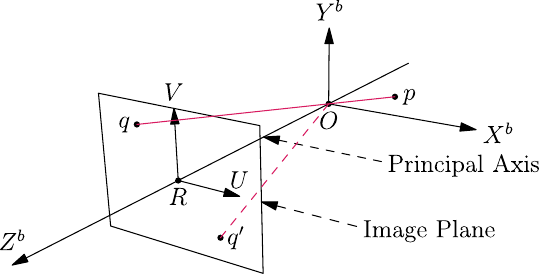}
  \caption{Pinhole camera geometry used to map image-plane coordinates to sky-ray directions. The focal length $f$ determines the relationship between each sensor location and its corresponding ray in the camera body frame.}
  \label{f:pinhole_model}
\end{figure}

Let $p$ be a direction defined by azimuth angle $\theta$ and polar angle $\phi$.
The azimuth angle is right-handed about the $Z^{b}$ axis taken from $X^{b}$.
The polar angle is taken from $Z^{b}$ towards $-Z^{b}$ in the plane defined by $Z^{b}$ and $p$.
The polar angle has the range $\ang{90} < \phi \leq \ang{180}$ so that it points away from the image plane.
Then, we define the forward projection in (\ref{e:projf}) and the inverse backward projection in (\ref{e:projb}).
The $\operatorname{atan2}$ function in (\ref{e:projb}) is similar to $\arctan$, but selects the correct quadrant for the resulting angle, whereas $\arctan$ only maps to the interval $[-\frac{\pi}{2}, \frac{\pi}{2}]$.

\begin{align} 
  \begin{bmatrix}
    u \\
    v
  \end{bmatrix}
  &= -f \tan(\theta)
  \begin{bmatrix}
    \cos(\phi) \\
    \sin(\phi)
  \end{bmatrix} \label{e:projf} \\
  \begin{bmatrix} 
    \theta \\
    \phi
  \end{bmatrix}
  &=
  \begin{bmatrix}
    \operatorname{atan2}(v, u)\\
    \operatorname{atan2}\left(\sqrt{u^2 + v^2}, -f\right)
  \end{bmatrix} \label{e:projb}
\end{align}

Once we have traced the the pixel coordinate $q'$ through the lens into a direction $p^b$ using (\ref{e:projb}), we want to find the sky point viewed along $p^b$.
We apply the transform between the $b$-frame and the $n$-frame, denoted $C^n_b$, which is determined by the orientation of the present state estimate.

\subsection{Measurement with Polarization Camera}

We measure the state of polarization at different sky points using a \ac{dofp} camera.
These devices capture polarization information in real-time by integrating a layer of micro-polarizer filters directly onto the pixels of an image sensor.
In the common configuration, a repeating pattern of 2x2 micro-polarizer arrays is superimposed on the pixel matrix.
We call each instance of the micro-polarizer array on the image sensor a \emph{superpixel}.
Each superpixel measures the intensity at orthogonal, linear polarization angles.
From such intensity measurements, the Stokes vector, thus the \ac{aop} and \ac{dop}, is derived.

\begin{figure}[t]
  \centering
  \includegraphics[width=2.75in]{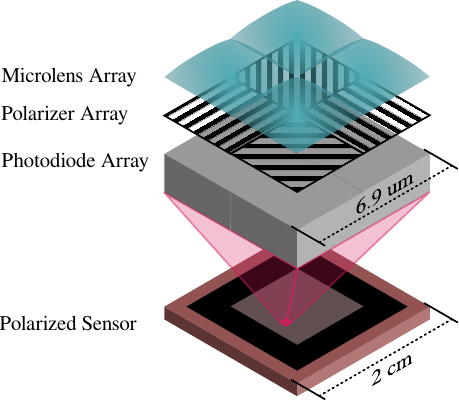}
  \caption{Division-of-focal-plane polarimeter superpixel structure. Four micro-polarizers measure intensities at orthogonal linear polarization angles, which are used to reconstruct the Stokes vector, \ac{aop}, and \ac{dop}.}
  \label{f:dofp}
\end{figure}

Fig.~\ref{f:dofp} depicts the structure of a single superpixel.
Incident light is bent by the microlens layer such that it becomes normal to the plane of the sensor.
Then, it passes through the micro-polarizer layer, which modulates the intensity of the incident ray by its \ac{aop} and \ac{dop}.
This modulated intensity is then measured by the photodiode layer.
We use the measurements from the photodiode layer to reconstruct the state of linear polarization of the incident ray via the Stokes vector.

The Stokes vector is derived from the intensity measurements for a single superpixel following~(\ref{equ:stokes}).
The intensity measurements are taken at \qty{0}{\degree}, \qty{45}{\degree}, \qty{90}{\degree}, and \qty{135}{\degree}.
These measurements are labelled $I_{0}$, $I_{45}$, $I_{90}$, and $I_{135}$, respectively.
The circular component of the Stokes vector, $S_3$, is undefined in this measurement scheme since we only capture linear polarization angles.
With the Stokes vector derivation, we can compute the \ac{aop} and \ac{dop} of the ray following (\ref{equ:aop}) and (\ref{equ:dop}).
We repeat this process for each superpixel to compute the polarization state for a full image.

\begin{gather} \label{equ:stokes}
  \begin{bmatrix}
  S_0\\
  S_1\\
  S_2\\
  S_3
  \end{bmatrix}
  =
  \begin{bmatrix}
  (I_{0} + I_{90} + I_{45} + I_{135}) / 2 \\
  I_{0} - I_{90} \\
  I_{45} - I_{135} \\
  undefined
  \end{bmatrix}
\end{gather}

\begin{align}
  \label{equ:aop}
  a^s &= 1/2 arctan (S_2 / S_1) \\
  \label{equ:dop}
  d &= 1/S_0 \sqrt{S_1^2 + S_2^2} 
\end{align}

We note that the \ac{aop} measured by a \ac{dofp} polarimeter is taken implicitly with reference to the \qty{0}{\degree} polarizer in each superpixel.
We call this the superpixel coordinate system and denote it with the $s$-frame.
However, our sky model defines the \ac{aop} in the ray coordinate frame.
As such, we must develop the transformation between the $s$-frame and the $r$-frame.

The transform from the sensor frame to the ray frame corresponds to a rotation about the sensor's normal vector.
The rotation is defined by the angle between the $X^r$ axis of the $r$-frame and the \qty{0}{\degree} polarizer orientation, which is aligned with the $X^s$ axis of the $s$-frame.
While all superpixels in the sensor frame share a common $X^s$ direction, the $X^r$ direction of the $r$-frame depends on the azimuthal orientation of the ray.
Since different rays propagate with different azimuthal orientations, the reference direction in the ray frame must be defined for each ray individually.

Considering a single ray shown in Fig.~\ref{f:scatter_event}, we develop the procedure for transforming the \ac{aop} reference frame.
Given the constant definition of the zenith vector, $U^n$, in the $n$-frame, we transform it into the $b$-frame by (\ref{e:u_from_n_to_b}) using the orientation of the camera to determine $C^b_n$.
Then, we can find the pixel coordinate of $U^b$ by representing it as an azimuth and polar angle and applying (\ref{e:projf}).
This process identifies the pixel on the image sensor that observes the zenith location.

\begin{equation} \label{e:u_from_n_to_b}
  U^b = C^b_n U^n
\end{equation}

Given the pixel location of the zenith vector, $(s, t)$, and the position of the $i^{th}$ pixel, $(x_i, y_i)$, we can approximate the orientation of the $X^r_i$ vector in the $s$-frame.
This vector is the reference direction of the \ac{aop} in the $r$-frame for a pixel of interest, $i$.
By specifying $X^r$ in the $s$-frame, we can find the corrective angle between the $X^s$ and $X^r$ vectors, and use it to determine the \ac{aop} in the $r$-frame.
Then, (\ref{equ:theta}) approximates the angle $\theta_i$ from the $X^s_i$ to the $X^r_i$ axis with some error $\epsilon_i$.
Given the angle $\theta_i$, we can transform the measured \ac{aop} from the $s$-frame to the $r$-frame by (\ref{equ:aop_s_to_r}).

\begin{align}
  \label{equ:theta}
  \theta_i + \epsilon_i &= arctan \frac{t - y_i}{s - x_i}\\
  \label{equ:aop_s_to_r}
  a^{r}_i &= a^{s}_i + \theta_i
\end{align}

The error term $\epsilon_i$ comes from simulating only the optical path of the zenith, rather than the optical paths for every pixel position.
When the orientation of the image sensor is aligned with the horizon, which we call the \emph{level} case, we find that $\epsilon_i = 0$ for all pixels i.e., our approximation in (\ref{equ:theta}) is exact.
However, when the image sensor is not aligned with the horizon, which we call the \emph{tilt} case, we find that $|\epsilon_i| > 0$ and increases with the magnitude of the tilt.
By applying this approximation, computing the transform is faster at the cost of reduced accuracy.

The fact that using the zenith position can provide an exact transform in the level case without the need for exhaustive simulation motivates its use in (\ref{equ:theta}).
In the tilt case, the geometry of the projection onto the image plane causes the approximation to lose accuracy.
If we did simulate the optical paths for every pixel, we could trace the $r$-frame through each transform the camera applies, ending with the exact angle from $X^s_i$ and $X^r_i$.
We use this approach to determine the behaviour of $\epsilon_i$ in the approximation.
We find that the error in such an approach is resonably small under the conditions of our application domain.

\begin{figure}[t]
  \centering
    \subfloat[Level case: roll~=~\ang{0}, pitch~=~\ang{0}, yaw~=~\ang{0}]{%
      \includegraphics[width=1.7in]{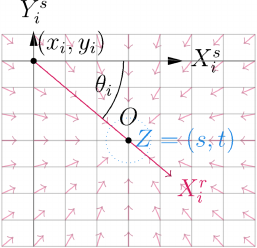}%
      \label{f:aop_level}%
    }\hfil
    \subfloat[Tilt case: roll~=~\ang{15}, pitch~=~\ang{-15}, yaw~=~\ang{15}]{%
      \includegraphics[width=1.7in]{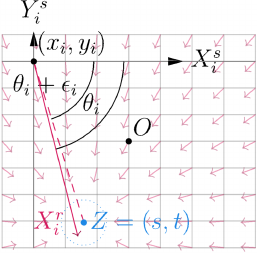}%
      \label{f:aop_tilt}%
    }\\
  \caption{Relationship between the ray-frame $X^r_i$ axis and the superpixel-frame $X^s_i$ axis for the $i^{th}$ pixel under level and tilted camera orientations. The angle between these axes defines the rotation required to compare measured sensor-frame \ac{aop} values with Rayleigh-model predictions in the ray frame.}
  \label{f:aop_transform}
\end{figure}

Fig.~\ref{f:aop_transform} compares the level case, where the $X^r$ axis points towards the zenith pixel, to the tilt case, where the $X^r$ axis is misaligned by the error term $\epsilon$.
We consider the $i^{th}$ pixel as an example of the behaviour of $\epsilon_i$.
We show the geometric interpretation of (\ref{equ:theta}) for this pixel.
In the level case, the error term is zero, since the $X^r$ axis aligns with the direction to the zenith pixel.
In the tilt case, we must include this error term, as the $X^r$ axis does not perfectly align with the direction to the zenith pixel.
The approximated orientation of the $X^r$ axis using (\ref{equ:theta}) is shown as a dotted line.

Our analysis of the error characteristics of the approximation lead us to believe it is a reasonable trade-off for the reduced computational complexity.
While there are regions of high $\epsilon$ in an image under the tilt case, they are limited to a fraction of the total image.
Many of the pixels in the image have correspondly small values of $\epsilon$.
Further, the mean absolute value of $\epsilon$ is relatively small, even for large pitch and roll angles.
In a typical road vehicle scenario, the pitch and roll values do not exceed \ang{5}, which limits the mean absolute value of $\epsilon$ to less than a degree.
In application domains where tilt is more common, researchers should consider alternative methods for transforming the \ac{aop}.

%% file: sections/dataset_instrumentation.tex
\subsection{Instrumentation}

\begin{figure}[t!]
    \centering
    \begin{tikzpicture}[
      dotpointer/.style={-{Circle[scale=0.8]}, ultra thick},
      labelbox/.style={
        fill=white, 
        ultra thick,
        rounded corners=3pt,       
        text width=2.5cm,          
        minimum height=0.3cm,      
        align=left,                
        font=\scriptsize, 
        anchor=south west
      }
    ]
        \node[anchor=south west, inner sep=0] (image) at (0,0) 
            {\includegraphics[width=0.6\columnwidth]{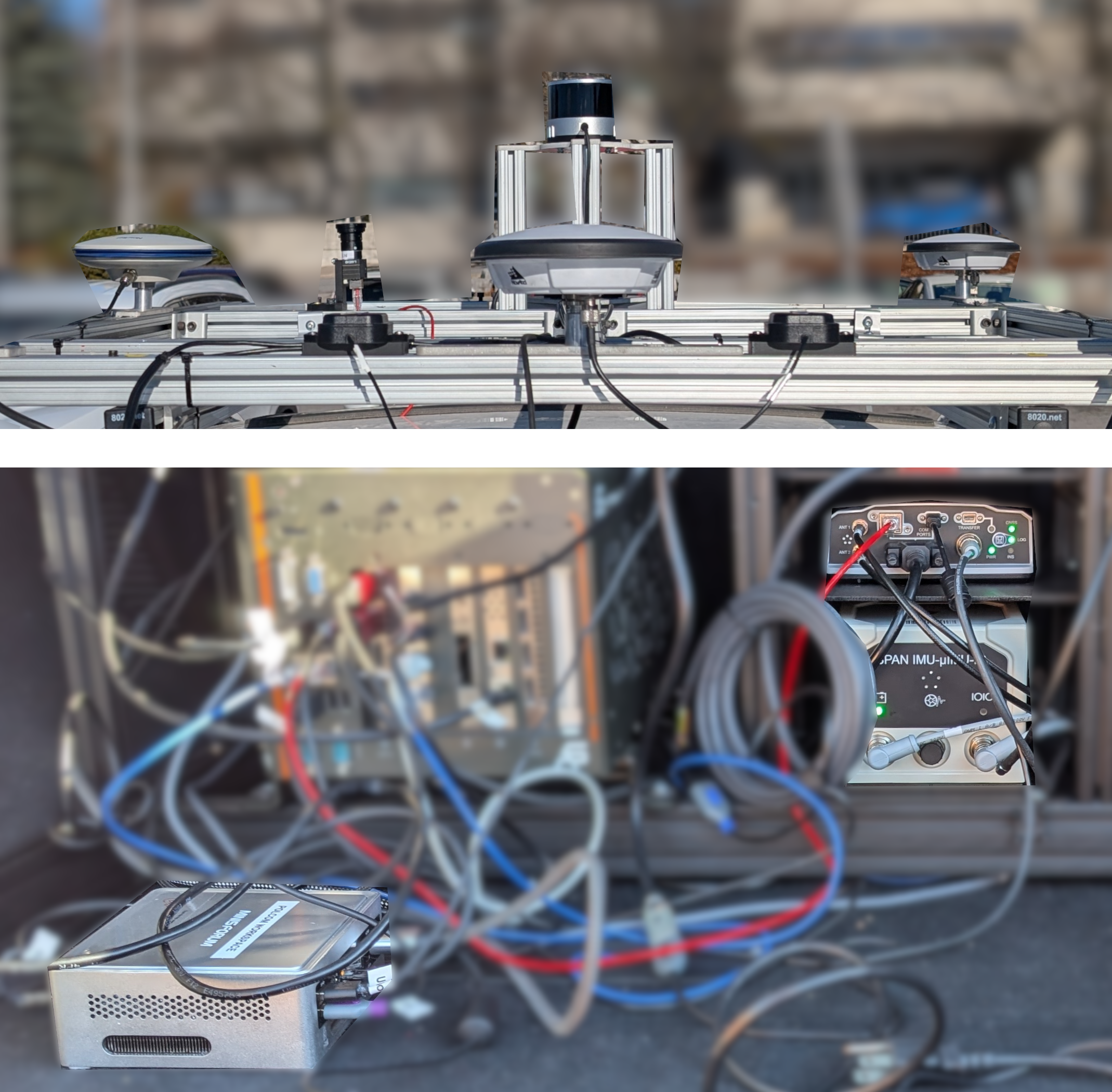}};
        
        \begin{scope}[x={(image.south east)}, y={(image.north west)}]

            \node[labelbox, draw=primary] (l_camera)   at (-0.6, 0.6) {Polarization Camera};
            \node[labelbox, draw=secondary] (l_antenna)  at (-0.6, 0.5) {Dual GNSS Antennas};
            \node[labelbox, draw=secondary] (l_pwrpak)   at (-0.6, 0.4) {PwrPak7};
            \node[labelbox, draw=secondary] (l_imu)      at (-0.6, 0.3) {IMU-µIMU-IC};
            \node[labelbox, draw=tertiary] (l_recorder) at (-0.6, 0.2) {Data Recorder};

            \node[labelbox, anchor = north east, text width=0.75cm, align=center] (l_roof)  at (0.95, 0.95) {Roof};
            \node[labelbox, anchor = south east, text width=0.75cm, align=center] (l_trunk) at (0.95, 0.05) {Trunk};

            \node[anchor=south west, draw=primary, thick, inner sep=0] (inset) at (0.05, 0.6) 
                 {\includegraphics[width=0.1\columnwidth]{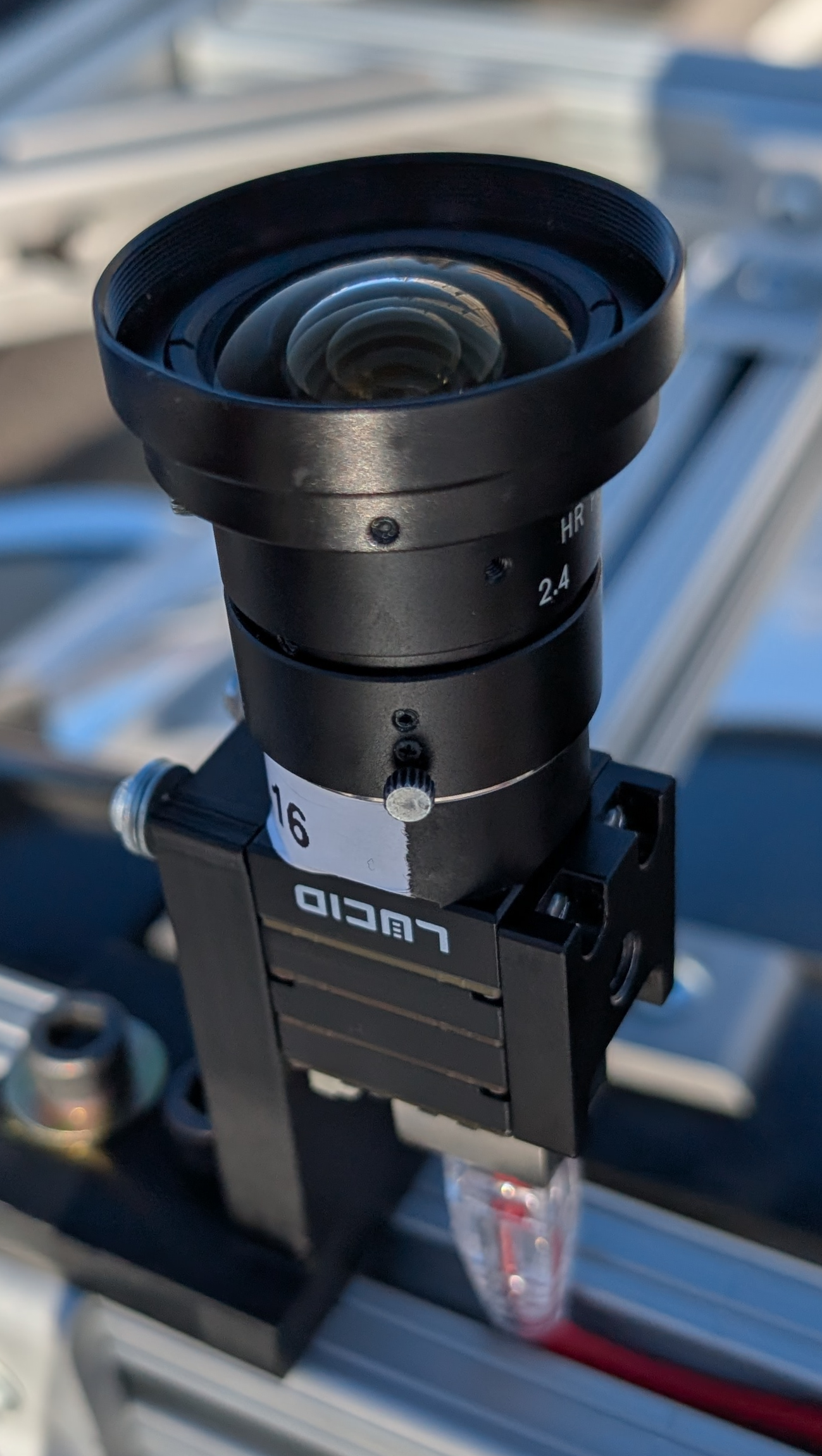}};
            \draw[primary, thick]         (0.28, 0.70) rectangle (0.36, 0.82);
            \draw[primary, thick, dashed] (0.36, 0.70) -- (inset.south east);
            \draw[primary, thick, dashed] (0.36, 0.82) -- (inset.north east);

            \draw[dotpointer, primary] (l_camera.east) -| (0.14, 0.70);
            \draw[dotpointer, secondary] (l_antenna.east) -| (0.50, 0.75);
            \draw[dotpointer, secondary] (l_antenna.east) -| (0.86, 0.75);
            \draw[dotpointer, secondary] (l_pwrpak.east) -| (0.80, 0.50);
            \draw[dotpointer, secondary] (l_imu.east) -| (0.80, 0.40);
            \draw[dotpointer, tertiary] (l_recorder.east) -| (0.20, 0.10);
        \end{scope}
    \end{tikzpicture}
    \caption{%
    Diagram of the experimental instrumentation we use to record the dataset.
    The polarization camera is highlighted with a close-up image for clarity.
    Instruments that are part of the reference system are distinguished using the yellow color.
    }
    \label{fig:setup}
\end{figure}
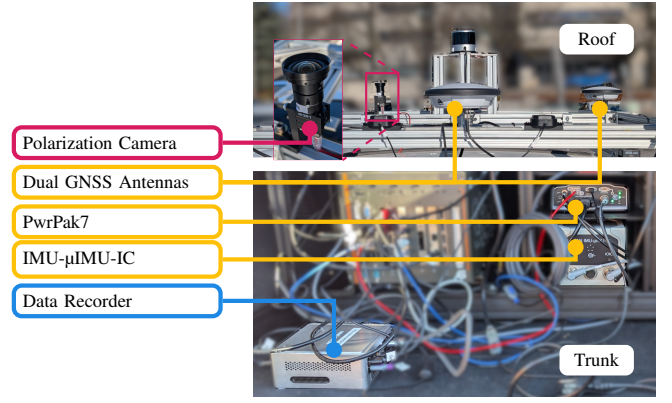

Our dataset is constructed using a modified consumer road vehicle, shown in Fig.~\ref{fig:setup}.
The modifications to the road vehicle are made to support advanced multi-sensor navigation research.
The on-board reference system has been applied to previous navigation research by Araujo et al.~\cite{araujo2025NavInst}.
The reference navigation solution is generated using a NovAtel PwrPak7 integrated system, configured with dual GNSS antennas and a high-performance NovAtel IMU-µIMU-IC.
The dual-antenna setup enables robust and accurate heading estimation, while the tightly coupled GNSS/INS integration provides reliable pose estimation under dynamic conditions.
To further enhance positioning accuracy, TerraStar Pro corrections are applied, enabling \ac{ppp} with centimeter-level accuracy.
The system outputs a high-rate, high-precision navigation solution at \qty{50}{\hertz}, providing position, velocity, and orientation estimates that serve as ground truth for performance evaluation.
In this work, we integrate a LUCID Vision Labs PHX050S-P/Q polarization camera to measure the skylight polarization pattern.
The camera is fixed to a rack on the roof of the car with the optical axis pointing to the zenith.
It is configured to capture a \num{2048} by \num{2448} image every \qty{200}{\milli\second}.
The camera has a \qty{3.5}{\milli\metre} focal length lens with a field of view of \ang{120} attached.

We fuse the measurements from the camera and the reference system using a nearest neighbour approach.
Each image is labelled with the closest time-wise measurement from the reference system.
The maximum time difference between an image and its label is \qty{50}{\milli\second}.
We do not need a more robust fusion strategy in this case since the polling rate of the reference system is much faster than the camera.
There is always a measurement from the reference system that is within the maximum time difference from every image.

\subsection{Data Collection}

We create a dataset that we use to evaluate the performance of the model optimization algorithm.
The dataset is comprised of three trajectories taken in Kingston, Ontario, Canada.
The trajectories were recorded sequentially on November 24, 2025.
Trajectory~1 started at \DTMdisplaytime{11}{37}{35} EST with a duration of \qty{10.10}{\minute}.
Trajectory~2 started at \DTMdisplaytime{12}{08}{49} EST with a duration of \qty{18.92}{\minute}.
Trajectory~3 started at \DTMdisplaytime{12}{46}{30} EST with a duration of \qty{13.93}{\minute}.
Times are reported in the local time zone.

\begin{figure*}[!t]
    \centering
    \subfloat[Trajectory 1]{\includegraphics[width=2.33in]{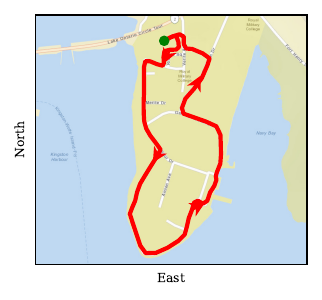}
    \label{fig:rmct}}
    \hfil
    \subfloat[Trajectory 2]{\includegraphics[width=2.33in]{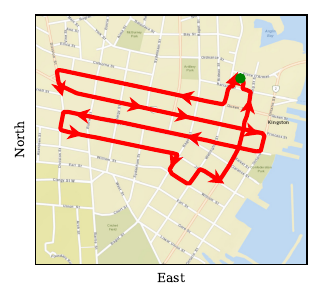}
    \label{fig:ur1t}}
    \hfil
    \subfloat[Trajectory 3]{\includegraphics[width=2.33in]{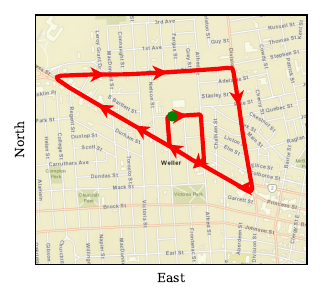}
    \label{fig:ur4t}}
    \caption{Urban road-vehicle trajectories used for experimental evaluation in Kingston, Ontario. Red points denote the reference-system position measurements, arrows indicate the direction of travel, and green markers identify the start and end locations. Basemap tiles attribution: Esri World Street Map.}
    \label{fig:traj}
\end{figure*}

The trajectories are shown in Fig.~\ref{fig:traj}.
Each position measurement provided by the reference system is plotted as a point on the red line.
The background map shows the north direction aligned with the y axis.
We show the direction of travel with arrows positioned along the trajectory.
The start and end position of each trajectory is shown with a green dot.

We now consider the relevant environmental conditions of the dataset.
The sky conditions are sunny with few passing clouds throughout all trajectories.
The altitude angle of the sun approximately varies from \qtyrange{23.27}{25.11}{\degree}.
Our trajectories pass through downtown Kingston, so there are intermittent obstrutions by buildings and tree foliage.
To establish the effect of these occlusions on the available skylight polarization information, we determine the percentage of available sky pixels.
We do not use this classification scheme when computing heading estimates via model optimization.

Let $S_0$ represent the intensity and $d$ represent the \ac{dop} of a candidate pixel.
We mark the candidate pixel as \emph{available} if it satisfies (\ref{e:available_intensity}) and (\ref{e:available_degree}) for some threshold intensity, $A$, and \ac{dop}, $B$.
In our application, we determined $A = 100$ and $B = 0.2$ by visually tuning the available regions using the image data itself.
The scale of the $A$ threshold is non-trivial since it borrows from the scale of the monocolor intensity pixels recorded by the polarization camera.
The $S_0$ parameter falls on the range $[0, 510]$, so we can consider the realized value of $A$ to represent the darkest \qty{20}{\percent} of possible pixel intensities.
Pixels which observe the sky during daytime rarely have an intensity lower than $A$, hence they are marked available.
The scale of the $B$ threshold is well-defined since the range of the \ac{dop} is always $[0, 1]$.

\begin{align} 
  S_0 > A \label{e:available_intensity} \\
  d > B \label{e:available_degree}
\end{align}

\begin{figure}[!t]
  \centering
  \includegraphics[width=3.5in]{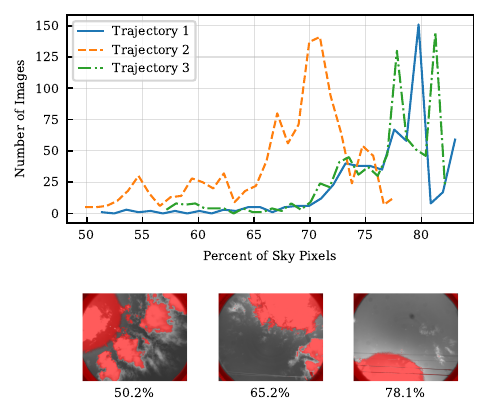}
  \caption{Distribution of available sky pixels for each trajectory. Example polarimetric images illustrate different levels of sky visibility, with red overlays indicating pixels classified as obstruction.}
  \label{fig:sky_px_distr}
\end{figure}

Fig.~\ref{fig:sky_px_distr} shows the distribution of sky pixels for each trajectory.
For each image in a trajectory, we determine the percentage of pixels which likely view the sky.
We show three examples of sky images with increasing percentages of sky pixels.
Pixels classified as obstruction are colored with a red overlay.
Trajectory~1 contains periods of cover by tree foliage, but this is not fully captured by the classification scheme.
Trajectory~2 has the lowest percentage of sky pixels due to the buildings present on either side of the roadway.
Trajectory~3 contains some large buildings, but it mostly passes through residential areas where the buildings only cover a small portion of the sky.

%% file: sections/results.tex
The performance of the model optimization method is summarized in Table~\ref{tab:results}.
We include the performance of Lu's method~\cite{lu2015HoughTransform} and Wang's method~\cite{wang2017BionicOrientation} for comparison.
We provide the RMS, P95, and maximum absolute error as measures of algorithm performance.
The P95 error represents the lowest error that contains \qty{95}{\percent} of all estimates.
Following, we explain our choices for the model optimization algorithm's hyperparameters.
Then, we discuss the merits of our algorithm based on our results in the context of the existing methods.

\begin{table}[t]
\centering
\caption{Summary of error in yaw estimate}
\label{tab:results}
\begin{tabular}{@{}llrrr@{}}
\toprule
\multicolumn{2}{c}{Test Case}           & \multicolumn{3}{c}{Yaw Error (\degree)}                   \\ \midrule
Method                                  & Trajectory   & RMS           & P95           & Max. Abs.  \\ \midrule
Lu's \cite{lu2015HoughTransform}        & Trajectory 1 & 2.930         & 5.279         & 20.415              \\
                                        & Trajectory 2 & 4.357         & 7.301         & 27.458              \\
                                        & Trajectory 3 & 5.688         & 8.597         & 20.059              \\ \midrule
Wang's \cite{wang2017BionicOrientation} & Trajectory 1 & 1.787         & 2.980         & 11.049              \\
                                        & Trajectory 2 & 2.673         & 4.483         &  7.162              \\
                                        & Trajectory 3 & 1.211         & 2.553         &  5.433              \\ \midrule
Proposed                                & Trajectory 1 & 0.889         & 1.855         &  3.600              \\
                                        & Trajectory 2 & 0.718         & 1.400         &  2.700              \\
                                        & Trajectory 3 & 0.760         & 1.500         &  4.700              \\ \bottomrule
\end{tabular}
\end{table}

\subsection{Hyperparameter Selection}

We optimize the objective function using a grid search over possible yaw angles.
We define the search space relative to the reference system to avoid checking angles far from the true yaw angle.
The bounds of the search space are \qty{+-5}{\degree} from the reference system's current yaw.
Within those bounds, the grid search has a resolution of \qty{0.1}{\degree}.
To generate an estimate, we evaluate the objective function at each grid location, and select the grid location that corresponds to its minima.
In practice, a fused solution may dynamically adjust the bounds of the search space to conform with the current uncertainty in its state.
For our purpose, we fix the bounds across all estimates.

We implement the camera model by loose matching to the optical characteristics of our real polarimeter.
We set the focal length of the camera model to \qty{8.0}{\milli\metre}.
We adjust the sensor dimensions to match the size of a superpixel on the real polarimeter.
The real polarimeter has a pixel size of \qty{3.45}{\micro\metre} in a 2048x2448 array.
These dimensions correspond to a superpixel size \qty{6.9}{\micro\metre} in a 1024x1224 array.
In the camera model we use these dimensions as we do not simulate the micro-polarizer array.
Notably, our algorithm achieves a high performance without any tuning of the camera model parameters.
We do not correct for any intrinsics of the real optical system.
This suggests that our approach may generalize to other measurement methodologies well.

\subsection{Merits of Proposed Method}

\begin{figure*}[!t]
  \centering
  \includegraphics[width=7in]{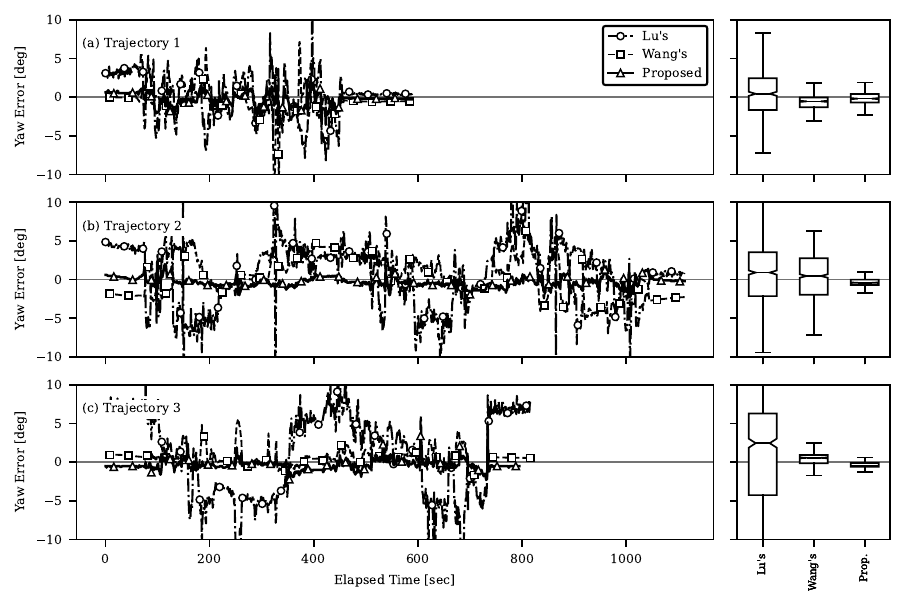}
  \caption{Yaw-estimation error for Lu's method, Wang's method, and the proposed full-image model-optimization method across the three vehicle trajectories. Time-series plots show frame-by-frame error, and box plots summarize each method's error distribution.}
  \label{fig:err}
\end{figure*}

Fig.~\ref{fig:err} shows the error in the yaw estimate over time, and as a box plot for each method.
The y axis shows the yaw error in degrees.
We compute the yaw error as the difference in the estimated yaw angle and the reference yaw angle.
The x axis show the elapsed time of the trajectory in seconds.
The error dynamics of each method are shown in the box plot.
This plot shows the median error as an orange line with the vertical extents of the box showing the interquartile range.
The interquartile range is the interval between the first and third quartiles of the yaw estimate error.
The tails on either side of the box represent the minimum and maximum of the error, when ignoring outliers.

We find that the yaw error shown in Fig.~\ref{fig:err} exhibits a drift-free property.
Unlike inertial sensors, heading estimation by skylight polarization does not accumulate error over time.
Absolute heading from an inertial sensor requires double integration over time, which amplifies small measurement errors in the angular acceleration.
Since each frame of the trajectory measures a new image of the sky, there is minimal coupling between adjacent state estimations.
This property makes our method ideal for the application of heading constraints on fused sensor systems with \ac{imu}s.

\begin{figure}[!t]
  \centering
  \includegraphics[width=3.5in]{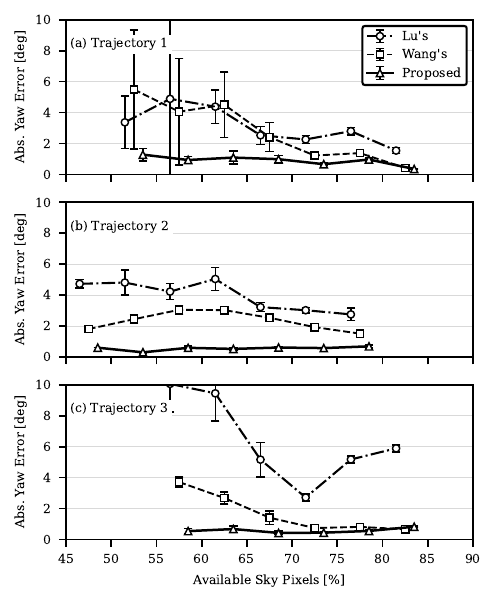}
  \caption{Absolute yaw error as a function of available sky pixels. Points show the mean error for each sky-visibility bin and error bars indicate the 95\% confidence interval, demonstrating the proposed method's robustness under partial sky occlusion.}
  \label{fig:err_vs_sky_pixels}
\end{figure}

We find that our method exhibits a low and stable absolute yaw error across varied environmental settings whereas the error properties of Lu's and Wang's method are generally coupled to the percentage of available sky pixels.
Fig.~\ref{fig:err_vs_sky_pixels} compares the yaw error as a function of the available sky pixels.
The x axis shows the available sky pixels as a percentage of the total pixels in an image.
The mean (y axis) and 95\% confidence interval (error bars) of the absolute yaw error are plotted for each trajectory.
In low available sky pixel settings, the existing methods exhibit larger error and variance when compared with the proposed method.
Further, their estimate variance decreases (i.e., the size of the error bars) as the percentage of available sky pixels increases, which indicates that such methods depend on low noise in the polarization measurement to produce confident estimates of yaw.
This analysis suggests that our method's \ac{dop} weighted objective function handles occluded conditions well.

Unlike many of the existing model optimization methods, where pixels can only be kept or ignored, our formulation of the objective function allows for strategic weighting of each pixel, which enables the estimation to use as much of the available polarization information as possible.
For example, Wang's method allows for certain e-vectors to be removed from the polarization data before the model optimization is performed~\cite{wang2017BionicOrientation}.
This property is leveraged by Fan and collaborators where they use the Rayleigh model to determine which pixels correspond to expected \ac{aop} values based on a binary threshold~\cite{fan2025OrientationRayleigh}.
By using a binary threshold, we meet the limitation of this approach, whereby we must select the value of this threshold.
Our method allows for a similar process, but instead of a binary threshold, we can introduce a weight factor to the objective function.
This additional granularity allows the estimation to rely on pixels with a large weight factor, while still considering pixels which may be excluded by a traditional binary threshold.

\subsection{Remaining Challenges with Proposed Method}

Many of the features proposed for this problem are symmetric.
This leads to an heading ambiguity problem where the solar meridian may be 180 degrees out of phase. 
Since our method does not rely on any single feature of the angle of polarization distribution, we are more robust against this case.
However, there is still a local minima at the antisolar meridian, which may cause challenges under high-noise environments.

The largest drawback of our method is its computational complexity.
Since we are performing a model optimization, we must evaluate the objective function at many different locations in the search space.
Each evaluation of the objective function produces a simulated image.
Simulating an image is not expensive in isolation, but simulating many images becomes expensive.
Due to our grid search approach to minimizing the objective function, we must check many states, which makes our method computationally expensive.
However, many of the states we check via grid search are far away from the true state, thus we do useless work.
There remains the potential to extend the model optimization approach with a more advanced search strategy, such as Riemannian gradient descent~\cite{bonnabel2013RiemannianGD}, which results in fewer iterations by only evaluating the objective function for \emph{promising} states, improving computational cost.
Further, in a practical fused sensor context, the polling rate of the absolute heading measurements from a polarization scheme does not need to be real-time, as the benefit of the polarization scheme is to constrain the time-coupled error bias inherent in inertial schemes, which already provide high polling rate.

%% file: sections/conclusion.tex
This paper presents a global model optimization framework for absolute heading estimation from skylight polarization using a vehicle-mounted \ac{dofp} polarimetric camera. 
The proposed approach combines a \ac{pcm}, a single-scattering Rayleigh sky model, and a solar ephermeris algorithm to synthesize the expected full-image \ac{aop} distribution for candidate vehicle states. 
By comparing the measured and simulated \ac{aop} fields with a \ac{dop}-weighted \ac{wrmse} objective function, the method avoids dependence on local features such as the solar meridian and uses the available polarization information across the observed sky. 
We also formulate the measurement-frame transformation required to compare \ac{dofp} measurements with ray-frame sky-model predictions and discuss the effect of non-horizontal camera tilt on this transformation.

Experimental evaluation on three urban road-vehicle trajectories demonstrate that the proposed method provides accurate, drift-free heading estimates under realistic sky occlusions from buildings, foliage, and clouds. 
Across the evaluated trajectories, the proposed method achieved sub-degree RMS yaw error, with RMS errors between \qty{0.718}{\degree} and \qty{0.889}{\degree}. 
Compared with representative feature-based and model-based polarization navigation methods, the proposed full-image optimization reduces heading error and exhibits lower sensitivity to the percentage of available sky pixels. 
These results suggest that skylight polarization can provide a useful passive and infrastructure-independent absolute heading constraint for ground-vehicle navigation, particularly as a complement to inertial navigation systems whose heading errors accumulate over time.

Several directions remain for future work. 
The current grid-search implementation is computationally expensive because it repeatedly evaluates full-image simulations over the heading search space. 
More efficient optimization strategies, including gradient-based or manifold search methods, could reduce the number of objective-function evaluations required for each estimate. 
The simulation and objective-function evaluation are also highly parallelizable at the pixel level, making implementation on GPU or embedded parallel hardware a promising path toward online operation. 
Future work should consider extending the optimization from heading-only estimation to full attitude estimation, investigate tighter integration with inertial and vehicle-motion constraints through filtering or sensor-fusion frameworks, and evaluate performance across broader atmospheric, seasonal, and geographic conditions.